\documentclass[default,iicol,sn-vancouver]{sn-jnl}% Default with double column layout

\usepackage{dblfloatfix}

\jyear{2023}%

\usepackage{xcolor}
\begin{document}

\title[Learn\&Drop]{Learn\&Drop: Fast Learning of CNNs based on Layer Dropping}


\author*[1]{\fnm{Giorgio} \sur{Cruciata}}\email{giorgio.cruciata@unipa.it}

\author[1]{\fnm{Luca} \sur{Cruciata}}\email{luca.cruciata@unipa.it}

\author[1]{\fnm{Liliana} \sur{Lo Presti}}\email{liliana.lopresti@unipa.it}

\author[2]{\fnm{Jan} \sur{Van Gemert}}\email{j.c.vangemert@tudelft.nl}

\author[1]{\fnm{Marco} \sur{La Cascia}}\email{marco.lacascia@unipa.it}

\affil*[1]{\orgdiv{Engineering Department}, \orgname{University of Palermo}, \orgaddress{\street{V.le delle Scienze, Ed. 6}, \city{Palermo}, \postcode{90128}, \country{Italy}}}

\affil[2]{\orgdiv{Computer Vision Lab}, \orgname{Delft University of Technology}, \orgaddress{\street{Van Mourik Broekmanweg 6}, \city{Delft}, \postcode{2628 XE}, \country{The Netherlands}}}


\abstract{
This paper proposes a new method to improve the training efficiency of deep convolutional neural networks. During training, the method evaluates scores to measure how much each layer's parameters change and whether the layer will continue learning or not. % expressed in a score. 
Based on these scores, the network is scaled down such that the number of parameters to be learned is reduced, yielding a speed up in training.
Unlike state-of-the-art methods that try to compress the network to be used in the inference phase or to limit the number of operations performed in the backpropagation phase, the proposed method is novel in that it focuses on reducing the number of operations performed by the network in the forward propagation during training.
The proposed training strategy has been validated on two widely used architecture families: VGG and ResNet. Experiments  on MNIST, CIFAR-10 and Imagenette show that, with the proposed method, the training time of the models is more than halved without significantly impacting accuracy. The FLOPs reduction in the forward propagation during training ranges from 17.83\% for VGG-11 to 83.74\% for ResNet-152.  These results demonstrate the effectiveness of the proposed technique in speeding up learning of CNNs. The technique will be especially useful in applications  where fine-tuning or online training of convolutional models is required, for instance because data arrive sequentially.%\jvg{Can you add a conclusion? ie: Answer the "so what?" question } \gc{The proposed method is  a promising solution for fine tuning CNNs on new domains look at their learning layers based on the proposed score}
}


\maketitle

\section{Introduction}

The use of deep learning models is increasing over time, since they perform well in various areas, such as computer vision, natural language processing, and speech recognition.
In computer vision, a popular deep learning model is the Convolutional Neural Network (CNN), which consists of deep models with many different layers processing the input by mapping it to the expected output \cite{lecun2015deep}.

The training time of CNNs is sometimes very long and could last hours, days or even weeks depending on the task, the size of the  dataset and the available hardware. Nowadays, the trend is to increase the depth and size of architectures~\cite{menghani2021efficient}; This usually leads to better performance despite a heavy computational cost.
Although modern deep CNNs are composed of a variety of layer types, convolutional layers are the building blocks of CNNs, and contribute greatly to the overall computational load of the network. 

In recent times, the deep learning community has focused on how to improve the efficiency of CNNs while saving time, computational costs, and energy without compromising the accuracy of the results~\cite{menghani2021efficient}. 
We can divide the methods that improve the efficiency of CNN into two categories~\cite{menghani2021efficient}, depending on the stage in which they seek to achieve better efficiency:
\begin{itemize}
\item methods for inference efficiency: this category includes methods that compress the network during the training phase to use a more compact model during inference~\cite{iurifrosio,choudhary2022heuristic,zemouri2020new,he2017channel,xu2021efficient};
\item methods for training efficiency: fall into this category methods that improve the efficiency of model training only, for example by freezing some layers during training but continuing to use the entire model during the inference phase \cite{xiao2019fast,liu2021efficient,zhang2019eager}.
\end{itemize}

At first glance, one might think that methods for inference efficiency are the most relevant because a model is trained once and used to infer multiple times. However, there are applications where new data arrives sequentially and the model has to be adapted and retrained using this new data. An example is visual tracking where the model is often used to represent the target appearance. Since appearance changes over time, it is required that the model adapt to such changes. In that case, slow retraining of the network may affect the real-time capabilities of the tracker. Also, recommendation systems or large language models (LLM) must be regularly retrained, and the training can last several weeks

Motivated by the above considerations, in this work we focus only on training efficiency by proposing to gradually compress the neural network during the training phase by dropping some convolutional layers and using the entire model (with all its layers) during the inference phase. We analyze the method on some of the most popular CNNs such as VGG~\cite{simonyan2014very} and ResNet~\cite{he2016deep} by observing their learning behaviour through gradient monitoring.

The two main steps in neural network training are forward and backward propagation; both steps have a high computational cost, which increases according to the complexity of the network.
In general, forward propagation is the process of passing data through the network from one layer to another. In each layer, the input data is processed taking into account the weight matrix of the layer itself and a suitable activation function to produce the layer output. The output of each layer becomes the input of the next layer repeating this process until the output of the final layer is produced.
The back-propagation algorithm  \cite{lecun2015deep} is used to train a neural network by updating its weights to minimize the error between the predicted and expected output. The algorithm computes the gradient of the loss function with respect to each weight via the chain rule, starting at the output layer and propagating the gradient backwards through the network from one layer to another. 

The magnitude of the gradient permits to understand how the parameters of the model vary during training: the closer it is an optimal point, the lower the value of the gradient. The variation of the gradient magnitude across the epochs provides a learning curve that can be calculated separately for each layer of the network. We have empirically found similarities among these learning curves when analyzing different CNNs and, in particular, we have found that the layers' parameters are learned sequentially from the first to the last layer. 
Thus, in this paper we propose to sequentially eliminate the convolutional layers during the training  phase. Layers to drop are selected based on a metric related to the layer  gradient. The adopted metric gives us an indication of which layer has stopped learning and can, therefore, be temporally eliminated from the model.
However, when we drop a layer, we have to find a way to feed the next layer to continue its training. This is done by feeding the remaining model (the one without the deleted layers) with the feature maps produced from the last dropped layer. 
We stress here that dropping the layers has a double consequence: the weights of the removed layers are no longer modified and, moreover, a compressed model is trained in the next epochs. During the test, the entire model is used and each layer will have as weights those obtained when the layer was selected for dropping.
This method shows an huge acceleration of the training process. We can summarize the main contributions of this work as it follows:
\begin{itemize}

\item A new method to improve the efficiency of training CNNs by dropping layers based on the metric derived from gradients.

\item A significant speed-up of the training process of CNNs compared to state-of-art methods, given by the possibility of processing directly feature maps extracted from dropped layers and calculating the back-propagation only for the remaining layers. We empirically show that, by using our technique, the training time of a CNN is more than halved.

\end{itemize}

The paper is organized as it follows. In Sec.~\ref{sec:rw}, we summarize the main differences between methods for inference efficiency and methods for training efficiency. We also detail how our method relates to former works. In Sec.~\ref{sec:dl}, we detail the way we score convolutional layers in a given architecture and how we select layers to be dropped. We illustrate how our approach can be used during the training process by highlighting its main steps. Sec.~\ref{sec:exp} presents the results of an extensive validation of our approach over different architectures belonging to the family of VGG and ResNet models demonstrating the effectiveness of the methods on CNNs of varying depth. Three different and well known, publicly available datasets have been used to demonstrate that the proposed approach does not affect much the accuracy values of the trained models and contributes to greatly reduce the training time. In particular, a discussion about the time reduction is presented in Sec.~\ref{sec:tr} while Sec.~\ref{sec:con} presents conclusions and future work.

%-------------------------------------------------------------------------

\section{Related Work}
\label{sec:rw}

In recent years, many works have focused on how to reduce the parameter size of deep learning models. This is quite a challenge considering the energy impact of training large networks.
The goal of compression techniques is to achieve a more efficient representation in a neural network, improving the generalization of the model if the model is overly parameterized. 
Model optimization is performed with respect to model size or training time in exchange for as little accuracy loss as possible.
A popular method of reducing the complexity of neural networks is to permanently remove neurons, filters, or layers for training. This technique is called pruning. 
Pruning can be performed based on different aspects of neural networks, for example it may be possible to remove parameters with low saliency scores from a pre-trained network. These techniques are often referred to as Optimal Brain Damage or Optimal Brain Surgeon, and were first introduced by LeCun et al. \cite{lecun1989optimal} and Hassibi et al. \cite{hassibi1993optimal}, respectively. The goal of this process is to minimize the impact of compressing the network on its performance, as measured by its validation loss. OBD approximates the saliency score by using a second-derivative of the parameters $(\frac{\partial^2 L}{\partial w^2_i})$, where $L$ is the loss function, and $w_i$ is the candidate parameter for removal. 
In \cite{luo2017thinet}, it is suggested that a greedy method be used to determine the minimum set of neurons needed to minimize the reconstruction loss, but this approach has a high computational cost.
Other methods that prune by ``saliency" consider the magnitude of the weights~\cite{han2015deep,han2015learning}.
The previously described methods for removing sparse neurons are examples of unstructured pruning methods. 
On the other hand, structural pruning aims at reducing the number of filters as in~\cite{veit2016residual} where the $L_1$-norm is used to select the filters to be removed without affecting the accuracy of the classification. A similar idea is presented in~\cite{polyak2015channel} where the feature map channels that do not contribute to the result are removed.

In addition to pruning filters and channels, there are also methods of pruning entire layers~\cite{chen2018shallowing,xu2020layer,elkerdawy2020filter}. Using different criteria, the selected layers from the network are removed to obtain a compact model. These methods claim the model obtained from layer pruning require less inference time and memory usage at runtime with similar accuracy values than the model obtained from filter pruning  methods. The work in \cite{chen2018shallowing} uses independently trained linear classifiers per layer to rank their importance. After ranking, they remove less important layers and fine-tune the remaining model. However, their method requires additional rank training. All the previously described methods fall into the category of inference efficiency methods because they produce compressed model to be used at inference time.

In other efficiency training approaches like in~\cite{tan2020accurate}, the gradient computation through the chain rule is stopped. 

To speed-up training and increase accuracy in very deep networks, AFNet~\cite{tan2020accurate} investigates a different use of back-propagation. 
Only a subset of layers are trained while the others are frozen. Frozen layers weights do not need to be updated during back-propagation.
In~\cite{xiao2019fast}, a metric $F$, named Freezing Rate, is defined as a function of the gradient values of the set of weights of a layer. This metric is used in~\cite{xiao2019fast} to decide which layer should be frozen during the training. The frozen layers must be subsequent to not break the layers chain, and therefore the freezing of the layers start from the first layer and advances to the subsequent ones. During training, the layers are not removed and therefore the computational advantage of the method concerns only limiting the calculation of the gradient and the modification of the weights of the frozen layers.
Our approach is inspired by~\cite{xiao2019fast}, in the sense that we adopt the same metric $F$. However, unlike the method in~\cite{xiao2019fast}, our method consists in performing a layer elimination (drop) during the training phase.  Once a layer is removed, the remaining model is fed through the feature maps produced by the last deleted layer. This speeds up both forward propagation, since there is no need to recalculate the feature maps of the deleted layers, and backward propagation, since the gradient is not calculated in the dropped layers and there is no need to update their weights. Our experiments demonstrate how this approach increases training efficiency by gradually reducing the number of FLOPs over the epochs.
At the same time our method is different from classic pruning methods, as our method does not permanently remove the layers by performing a network compression but only temporarily drops them during the training phase. In fact, in the test phase, the model  will contain all the layers of the one originally created.

%-------------------------------------------------------------------------

\begin{figure*}[t!]%
\centering
  \includegraphics[width=.49\linewidth]{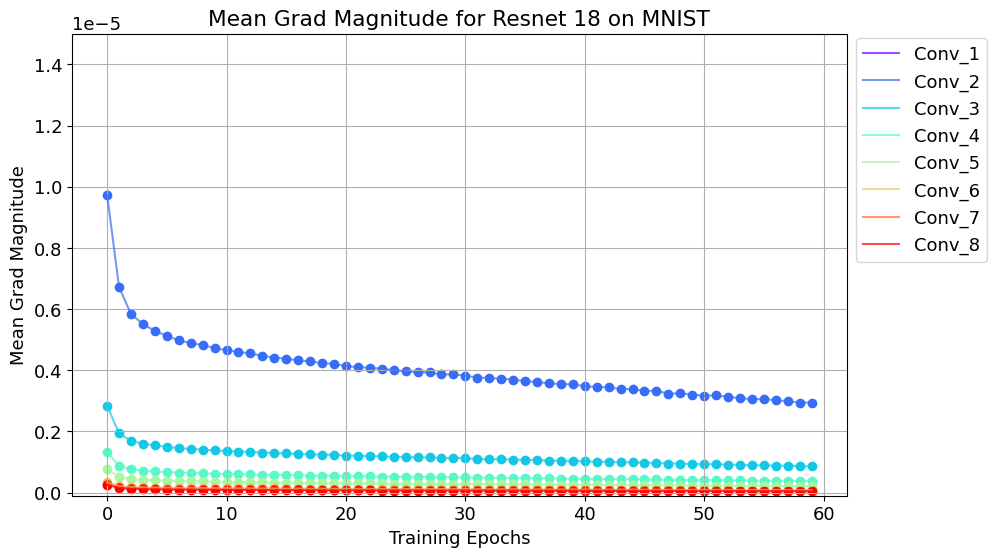}
  \includegraphics[width=.49\linewidth]{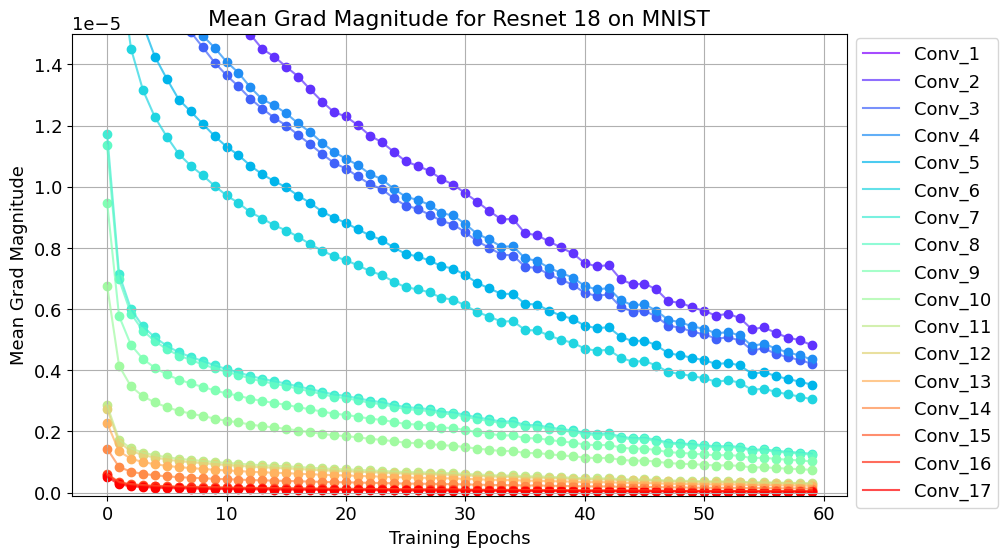}
  \caption{
  The two graphs represent the average absolute partial derivative value of each convolutional layer in the VGG-11 (left) and ResNet-18 (right) models on the MNIST dataset. The average absolute partial derivative value is indicated on the y-axis and the epoch on the x-axis. Each curve represents a different layer with colors from purple, for layers closest to the input, to red, for layers closest to the output. 
  The figure suggests that weights in the first layers undergo much higher changes than the weights in the layers closest to the output, especially at the beginning of the training. Thus, partial derivative values can help measuring if a layer is still learning or not. However, since the weights across the layers may have different magnitudes, we cannot use directly the average absolute partial derivatives. Instead, some form of normalization is needed.
  }
  \label{magnitude_gradients}
\end{figure*}

\section{Dropping Layers for Training Efficiency}\label{sec3}
\label{sec:dl}
Our method is described in Algorithm~\ref{algo1} and its main steps are:
\begin{itemize}
    \item Compute the ``Layer Importance metric" based on the gradient of the layer's weights;
    \item Apply the ``Fast Learning" algorithm, the core of our method. The algorithm consists of steps to: (1) select the layers to be dropped, (2) split the network into a ``tail", composed of the dropped layers, and a ``head", composed of the layers to still train, (3) compute and store output feature maps from the tail, (4) train the head by using output feature maps from the tail.
\end{itemize}

\subsection{Layer importance}
Our layer dropping method is based on the observation of the loss function gradients $\nabla g$. Generally, the gradient values can be interpreted as the rate of change of the weights. The sign of the partial derivatives represents instead the inverse direction in which weights should be changed to reach a minimum.

Given a neural network with layers $L = \{ l_0, l_1, ..., l_L \}$, the average absolute partial derivative value $ \overline{  g_l^{(k)} } $ corresponding to the weights of the $l$-th layer is calculated as: 
\begin{equation}
 \overline{  g_l^{(k)} } = \frac{1}{N}\sum_{i=1}^{N} \sum_{j=1}^{M} \mid g_{lij}^{(k)} \mid 
\label{eq1}
\end{equation}
where $N$ is the number of weights in layer $l$, $M$ is the number of iterations in epoch $k$, and $g^{(k)}_{lij}$ is the partial derivative of the loss function with respect to the $i$-th weight in layer $l$ at the $j$-th iteration in epoch $k$.
Fig.~\ref{magnitude_gradients} shows two graphs representing the average absolute partial derivative value $ \overline{  g_l^{(k)} } $ of each convolutional layer in the VGG11 and ResNet18 models. The layers closest to the input in both networks have a greater average partial derivative value than the layers closest to the output. In practice, the figure suggests that weights in the first layers undergo much higher changes than the weights in the layers closest to the output, especially at the beginning of the training. 

Thus, partial derivative values can help understanding if the weights of a layer are still changing or not. However, using directly the average absolute partial derivative values could be misleading since the weights may have small magnitude but still change. 

Hence, we decided to adopt the metric proposed in~\cite{xiao2019fast}, and define for the $l$-th layer the score:
\begin{equation}
P_l^{(k)} = 1 - \frac{\sum_{i=1}^{N} \mid \sum_{j=1}^{M}g_{lij}^{(k)} \mid}
{\sum_{i=1}^{N}\sum_{j=1}^{M}\mid g_{lij}^{(k)} \mid }
\label{eq2}
\end{equation}
with $0 \le P_l^{(k)} \le 1$, where $P_l^{(k)}$ measures the degree of changes of the weights in layer $l$ at the $k$-th epoch. 
$P_l^{(k)}$ will be $1$ if the partial derivatives cancel each other across the $M$ iterations. In such a case, within the epoch the layer weights do not change much and, intuitively, the layer has stopped to learn.
$P_l^{(k)}$ will tend to $0$ if most of the partial derivatives are in the same direction across iterations. In this case, layer weights are changing during the epoch. Thus, the layer is learning something about the problem to solve. We note here that the normalization factors make the scores comparable across the layers despite the different magnitude of the weight's partial derivatives.

Under this point of view, the score $P_l^{(k)}$ is measuring the importance of the layer during training. Layers with a score close to 0 must be trained. Layer with a score approaching 1 are not learning much and probably can be dropped  to speed up the model training.
Unlike~\cite{xiao2019fast}, where this score is used to freeze the layer and stop the back-propagation computation up to the $l$-th convolutional layer, our algorithm uses $P_l$ to drop the $l$-th convolutional layer. The feature maps  produced by the last dropped layer are used as input to the remaining model.
 
\subsection{Improving training efficiency}
In our approach, the removal of layers from the model to improve the training efficiency must take place in sequential order. At the $k$-th epoch, the layers to be dropped are selected based on the importance score $P_l^{(k)}$.

Thus, our fast learning algorithm works as it follows:
\begin{enumerate}
\item At the end of epoch $k$, the metric $P_l^{(k)}$ is calculated for each layer $l$. The score values are then standardized:
\begin{equation}
P_l' = \frac{P_l - \overline{P}}{\sigma_p}
\label{eq3}
\end{equation}
where $\overline{P}$ and $\sigma_p$ represent the average score over the layers and the standard deviation respectively. We omitted the apex $k$ for simplicity.

Ideally, we want to drop all subsequent layers for which the parameters do not change much anymore starting from the first layer of the network. 
The standardized scores $P_l'$ can have positive and negative values. Positive values indicate that the layer's weights are changing less than the average (hence the layer is likely not learning much), while negative values indicate that the layer's weights are changing more than the average (hence the layer is still learning something).

The problem of selecting the subsequent layers to drop starting from the first layer turns into the problem of finding the sub-vector of maximum sum starting from the first element of an array. In our case, the array represents the list of scores $P_l'$ with $l \in L$. 

Let us assume that the current layers in the model are $L=\{l_z, l_{z+1}, \dots, l_{L}\}$. Candidate layers to drop are $l_z...l_{n^*}$ with $n^*$ computed as:
\begin{multline} 
n^{*} = \min_{t} \{t \in [z, \ldots , L-1] : P_{l_t}' > 0 \\
\land P_{l_{t+1}}' < 0  \}.
\label{eq4}
\end{multline}

\item As soon the candidate layers to drop $l_z...l_{n^*}$ are found, to avoid dropping them too early, we estimate the median $M_c$ of the scores $P_{l_t}'$ with $l_t \in l_z...l_{n^*}$ (namely the scores of the candidate layers to drop), and compare it with the median $M_d$ of the scores $P_{l_t}'$ with $l_t \in l_0...l_{z-1}$ (namely the scores of the layers dropped in previous iterations and estimated when the decision of dropping the layers was taken). We perform layer dropping if $M_c\ge M_d$. In this way, we limit the effects that early layer dropping may have on the network accuracy value.

Once the layers to drop are identified, the network is split into two parts:  the ``tail", composed of the layers in the network up to $l_{n^*}$, and the ``head", composed of the layers from $l_{n^*+1}$ to the network output.

\item In epoch $k+1$, the tail is used to extract feature maps. These feature maps are stored on a memory, such as a disk, and are also used to feed the head to continue its training.

\item In epoch $k+2$, the stored feature maps are retrieved from the memory and used to train the head.
\end{enumerate}
These $4$ steps are within an iterative procedure repeated until the maximum number of epochs is reached or the convolutional layers are exhausted, namely the head does not have any layer.

Our approach differs from the one in~\cite{xiao2019fast}. In the latter approach, layers with a high score are not ``physically" removed from the network but their weights are not trained. The main limitation of the approach in~\cite{xiao2019fast} is that, during forward propagation, the data must be processed at each iteration even by layers for which the weights are not updated. Our approach overcomes this limitation by removing layers in order starting with the first. We experimentally demonstrate the advantages of this approach in significantly reducing the computational cost of the training process.

Furthermore, in~\cite{xiao2019fast} layers to exclude from the training are selected after a prefixed number of epochs based on the vector $f = [f_1,f_2, ..., f_n]$. Each value $f_i$ indicates the number of epochs between one freezing and another at a specific learning rate. 

Hyper-parameters in $f$ are empirically defined and change over the adopted datasets. In our approach, the decision to drop a layer is fully automatic. After each epoch, the method analyzes the scores $P_l^{(k)}$ to detect candidate layer to be dropped and, as already described, the decision to remove the layers or not depends also on the median of the estimated scores.

\begin{algorithm}[!ht]
	\caption{Fast Learning by layer dropping  }
	\label{algo1}
	\begin{algorithmic}[1]
            \Require
            \Statex $model$, a model of $L$ layers with randomly initialized parameters; 
            \Statex $e_1>0$, number of warm-up epochs; 
            \Statex $e_2>0$, number of training epochs;
            \Statex $L_0$, number of dense layers + 1;
            \Ensure
            \Statex Trained $model$;
            \State Initialize models $head$ and $tail$
            \State Initialize $Data$ with the training images 
            \State $save\_features =$ False.
            \State Train $model$ for $e_1$ epochs on $Data$
            \State $L=model$.layers.length()
            \State $head$.layers = $model$.layers$[0:L-1]$
            \State $tail$.layers = []
            \State $features\_maps = []$
            \State Initialize $P_l'$ with $l$ in $head$
		    \For {$k = 0$ \textbf{to} $e_2$} 
                \If {$save\_features$}
                    \State $save\_features$ = False
                    \State $features\_maps$ = $tail$($Data$)
                    \State Train $head$ on $features\_maps$ 
                    \State Update $model$ weights based on $head$
                    \State store $features\_maps$ to memory
                \Else
                    \If {$features\_maps != []$}
                        \State Initialize $Data$ with $features\_map$
                        \State $features\_maps = []$
                    \EndIf
                    \State Train $head$ for $1$ epoch on $Data$
                    \State Update $model$ weights based on $head$
                    \If {$L>L_0$}
                    \For {$\forall$ conv. layer $l$ in $head$} 
                    \State update $P_l'$ (Eqs.~\ref{eq2}\&~\ref{eq3})
                    \EndFor
                    \State Find $n^*$ (Eq.~\ref{eq4}) for layer dropping
                    \State Estimate median values $M_c$ and $M_d$
                    \If {$ n^* > 1 \land M_c \ge M_d$ }
                        \State $save\_features =$ True
                        \State $tail$.layers = $head$.layers$[0:n^*]$
                        \State $head$.layers.pop($0:n^*$)
                        \State $L=head$.layers.length()    
                   \EndIf
                   \EndIf
                \EndIf
                \State validate $model$ on validation set
		\EndFor
	\end{algorithmic}
\end{algorithm}

\subsection{Fast-Training Algorithm}
Our approach is described in Algorithm~\ref{algo1}. 
It starts with a few warm-up epochs $e_{1}$ where the $model$ is trained to move from the initial random weights. After this warm-up, the weights of $model$ are copied to the $head$ model, which is initially equal to $model$, while the $tail$ model is empty. 
From now on, only the $head$ model is trained. The $tail$ model stores the dropped layers and is used to estimate the features maps needed to feed the $head$ model. Each time the $head$ model is trained, the corresponding weights in the $model$ are updated accordingly. In practice, $model$ always contain all layers, whose weights are iteratively updated based on the weights learned by the $head$ model. 
The $save\_features$ flag is used to indicate whether the $tail$ model should be used to estimate feature maps using the dropped layers. 
$Data$ stores the data for training the model. Initially, $Data$ stores the training images. When layers are dropped, $Data$ stores the features maps produced by the dropped layers, i.e. the $tail$ model.
At each iteration, the layer importance $P_{l}'$ is recomputed only for each layer of the $head$ model as described in Equations~\ref{eq2} and~\ref{eq3}. 

Then, $n^*$ is  computed based on Eq.~\ref{eq4}. Layer dropping is performed if it is found a maximum-sum sub-sequence of scores $P_{l}'$ starting from the first layer of the $head$ that includes at least one layer and the median value of the scores in the found sub-sequence is greater than  the median score of the previously dropped layers. The scores of the dropped layer are not recomputed every time but stored during the training process and kept updated till the layer is not dropped.
The index $n^*$ is also used to further compress the $head$ model. In particular, the $tail$ model stores the dropped layers, namely the first $n^*$ layers of the $head$ model. These same layers are pulled out from the $head$ model, resulting in a reduced model.

The described process is iterated until only the last convolutional and dense layers remain; they continue to train till the maximal number of epochs $e_2$ is not reached. 

We emphasize that the whole model ($model$) is tested on a validation set; this proves that removing the layers does not affect the accuracy of the original model. 
The model is optimized using SGD method in order to maintain a fixed learning rate over the different iterations. This is not a limitation, and other optimizers might be used as well. Also, early stopping may be included in the algorithm to add more regularization. In our experiments, we did not use early stopping to compare different training strategies on equal terms of number of epochs.

 \begin{figure*}[t!]%
\centering
  \includegraphics[width=.49\linewidth]{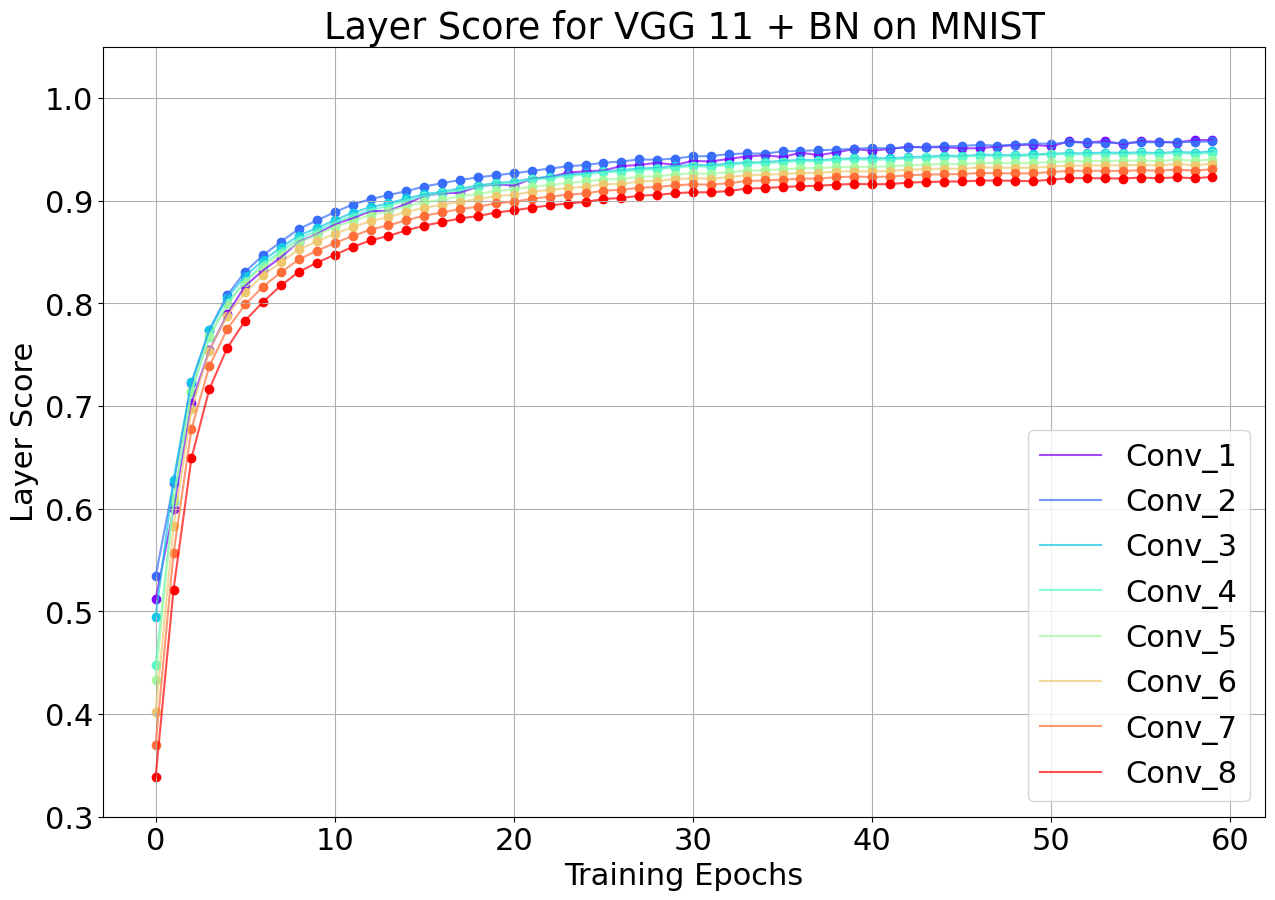}
  \includegraphics[width=.49\linewidth]{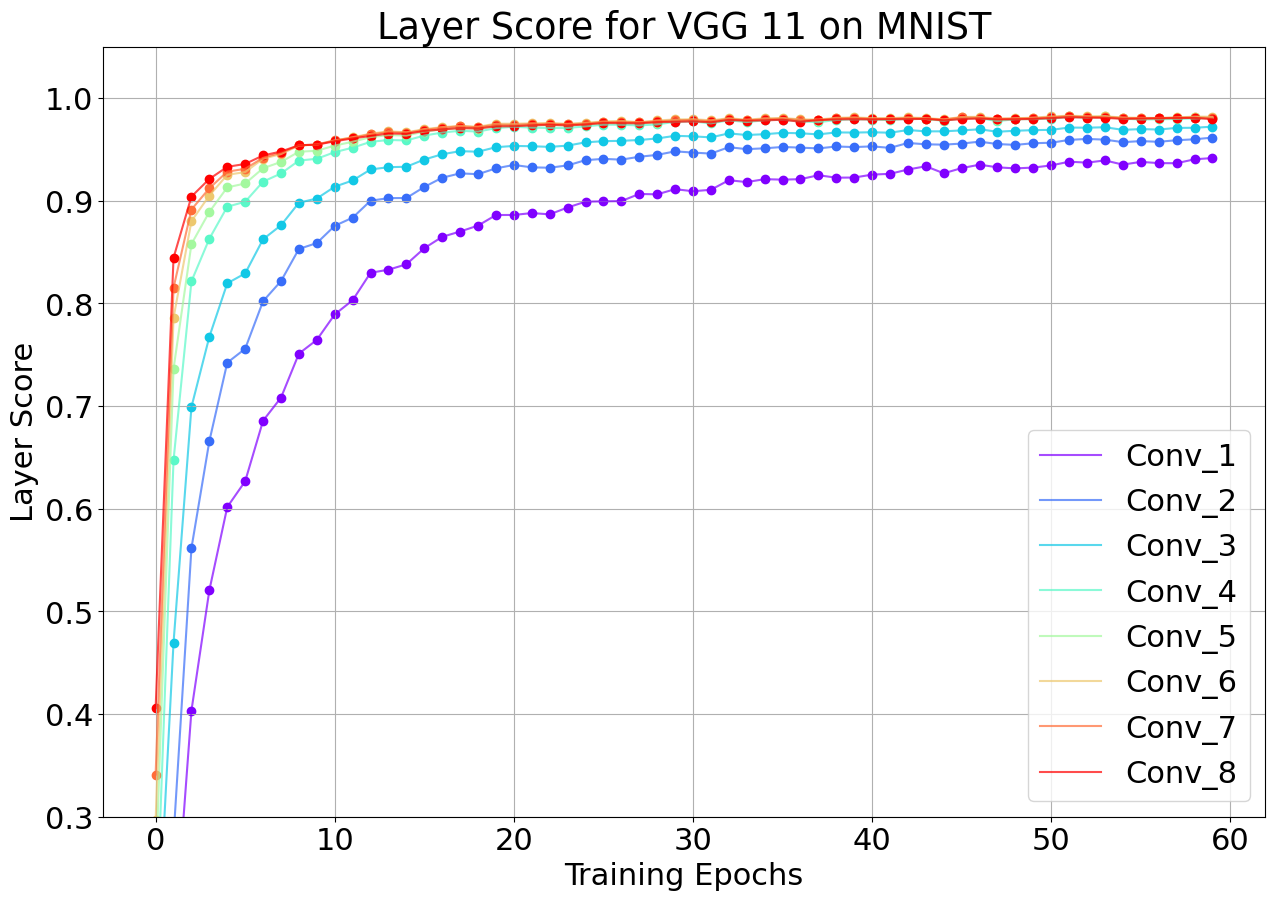}
  \caption{
  The two plots show the scores $P_l^{(k)}$ on the MNIST dataset for a VGG-11 trained with batch normalization (on the left) and without batch normalization (on the right). Adding a bath normalization reverses the order of the  score curves computed for each layer across the epochs. This is because batch normalization speeds up learning in neural networks by normalizing the inputs to each layer, which reduces the internal covariate shift. This makes the optimization more stable and allows the network to learn more quickly and with higher accuracy. As an effect, layers are learned sequentially from input to output. Thus, based on our experiments, to use our technique with a VGG it is recommended to include the batch normalization layer (one after each convolutional layer).
  }
  \label{fig:bachnormagraph}
\end{figure*}

\begin{figure}%
\centering
  \includegraphics[width=.99\linewidth,height = 0.20\textheight]{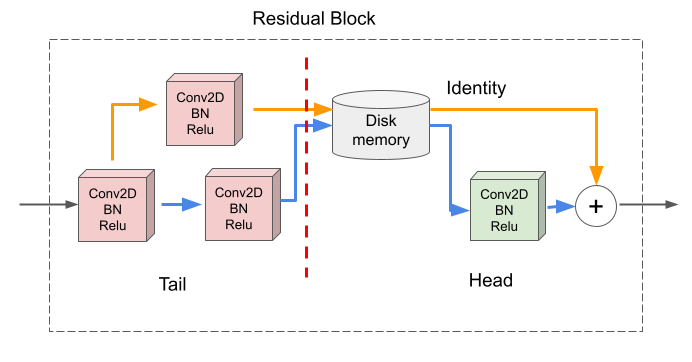}
  \caption{The image shows how dropping takes place in the ResNet at any residual block. The saved feature maps come from the layers in red, corresponding to the layers inside the residual block on the main track and on the skip connection. The layers on the left of the red dotted line are dropped and belong to the tail. The extracted feature maps are saved to the memory. From the next epoch, the head model (on the right of the red dotted line) will directly use the feature maps from the memory for training purposes.}
  \label{resnetblock}
\end{figure}

\section{Experimental Results}
\label{sec:exp}

In this section, we present results of the validation of our method.
We first present the selected neural architectures used to assess our method. We also provide details about how to apply our method to improve the training efficiency of these architectures. Then, we detail the datasets selected to perform the experiments and the hyper-parameters adopted on each dataset.

The results of our experiments are reported in Tables~\ref{tab:vgg} and~\ref{tab:resnet}.
In each table, ``Network" indicates the neural architecture; ``Dataset" specifies the dataset used for training and testing the model; ``SGD", ``Freezing" and ``Dropping" indicate the strategy used  to train the network. In particular, SGD is the baseline method, namely the model is trained in a standard way without attempting any training efficiency. Freezing represents our implementation of the method in~\cite{xiao2019fast} where, however, layers to freeze are selected in the same way as we do in our method. In this case, weights of selected layers are excluded from the training (frozen) but the layers are not physically removed from the trained model. The column Dropping reports results of our layer dropping method. 
The columns indicated with ``T" report the duration of the training and refers to the time in minutes required to complete the expected number of training epochs, including the warm up epochs. Columns ``A" report the test accuracy values, measuring the percentage of correct predictions made by the model on the test set. Finally, columns ``$\Delta T$" report the percentage of time saved by applying a training efficiency strategy (Freezing or Dropping) with respect to the baseline (SGD).
In particular, we compute this metric as:
\begin{equation}
    \Delta T = \frac{T_{SGD} - T}{T_{SGD}} \cdot 100.
\end{equation}
All experiments have been carried on a machine equipped with: GPU RTX 3090 24GB, RAM 96 GB, processor Intel(R) Xeon(R) CPU E5-2403  1.80GHz. 

In our implementation, feature maps produced by the dropped layers are stored on disk directly as PyTorch tensor using the ``Pickle" Python package~\cite{Pickle}, that implements binary protocols for serializing and de-serializing a Python object. We experimentally noted that using Pickle is faster than writing and reading files on disk with the Numpy package~\cite{numpy} and PyTorch~\cite{torch}.

\subsection{Neural Architectures}
In this paper we focused on improving training efficiency of CNNs. To assess our method we considered two neural network architectures widely adopted in the computer vision field. 
VGG (Visual Geometry Group) is a convolutional neural network introduced in~\cite{simonyan2014very}. The VGG architecture is characterized by its depth and the use of small convolutional filters. It consists of a sequence of convolutional layers, followed by a sequence of fully-connected layers. There are several configurations of the VGG architecture. The smaller version is the VGG-11 with only 11 layers. The VGG-16 has 16 layers and the VGG-19 has 19 layers. 
As the number of layers in VGG increases, so do the training time and memory requirements.

From the experiments carried out with VGG, we found out that the order of the curves representing the layer scores $P_l'$ across the epochs depends on the inclusion in the model of the batch normalization layer. In fact, as it can be seen in Fig.~\ref{fig:bachnormagraph}, we note that the scores order of the layers of VGG+BN is inverse respect to the scores order of the VGG without BN. This is because batch normalization speeds up learning in neural networks by normalizing the inputs to each layer, which reduces the internal covariate shift. This makes the optimization more stable and allows the network to learn more quickly and with higher accuracy. Hence, based on our experiments, to use our technique with a VGG it is recommended to include the batch normalization layer (one after each convolutional layer). 

ResNet (Residual Network) is a convolutional neural network introduced in~\cite{he2016deep}. ResNet is characterized by the use of residual blocks, which help to alleviate the vanishing gradient problem and allow for the creation of much deeper neural networks.
The original ResNet architecture has several configurations, including
ResNet-18, a relatively small version of the ResNet architecture with only 18 layers. ResNet-50, ResNet-101, and ResNet-152 are much deeper versions of the ResNet architecture with 50, 101, and 152 layers, respectively.

We point out that in the case of ResNet, we have to save not only the features maps but also the output of the skip connections to maintain the original behaviour as shown in Fig.~\ref{resnetblock}.

\begin{table*}[!t]
\small
\begin{center}
\begin{minipage}{\textwidth}
\caption{Fast Training of VGG architectures. SGD refers to the standard training strategy of the entire model. Freezing refers to excluding the parameters of some layers from the training without removing the layers from the model. Dropping is our method where layers are deleted from the trained model. T is the training time in minutes. A is the test accuracy value. $\Delta T$ is the percentage of reduced training time with respect to the time of SGD. 
 The table shows that our method allows a reduction of the training time of more than 58\% for all the VGGs trained on the different datasets. Best accuracy values (in bold) are achieved of course with SGD. Freezing or Dropping the layers achieve similar accuracy values, which are slightly inferior to the one obtained by SGD. However, in terms of training time reduction $\Delta T$, our method achieves the best results (in bold).
} \label{tab:vgg}
\begin{tabular*}{\textwidth}{@{\extracolsep{\fill}}llcccccccc@{\extracolsep{\fill}}}
\toprule%
& &\multicolumn{2}{c@{}@{}}{SGD} & \multicolumn{3}{@{}c@{}}{Freezing}& \multicolumn{3}{@{}c@{}}{Dropping (Ours)} \\
\cmidrule(lr){3-4}\cmidrule(lr){5-7}\cmidrule(lr){8-10}%
\textbf{Network} & \textbf{Dataset} & \textbf{T (min)} & \textbf{A (\%)} & \textbf{T (min)} & \textbf{A (\%)} &\textbf{$\Delta T$ (\%)}& \textbf{T (min)} & \textbf{A (\%)} &\textbf{$\Delta T$ (\%)} \\
\midrule
VGG-11 &MNIST & 20.83 & \textbf{98.64} & 19.58 & 98.25 & 6.00 & \textbf{8.74} & 98.25 &\textbf{58.04}\\
VGG-11 &CIFAR-10& 23.83 & \textbf{92.02} & 23.54 & 91.72 &1.21 &\textbf{ 8.21} & 91.72 &\textbf{65.54}\\
VGG-11 &Imagenette & 61.01 & \textbf{75.33} & 59.33 & 74.08 & 2.75& \textbf{18.32} & 74.08 &\textbf{69.97}\\
\hline
VGG-16 &MNIST  & 22.54 & \textbf{98.85} & 21.45 & 98.26 & 4,24 & \textbf{9.01} & 98.26 & \textbf{60.03}\\
VGG-16 &CIFAR-10& 26.54 & \textbf{93.12} & 24.94 & 92.84 &6.03& \textbf{9.56} & 92.84 &\textbf{63.98}\\
VGG-16 &Imagenette& 74.73 & \textbf{78.76} & 71.21 & 77.83 &4.71& \textbf{25.23} & 77.83 &\textbf{66.24}\\
\hline
VGG-19 &MNIST  & 23.02 & \textbf{98.52} & 22.68 & 96.22 &1.48&\textbf{9.45} & 96.22 &\textbf{58.95}\\
VGG-19 &CIFAR-10& 27.02 & \textbf{93.10} & 25.78 & 91.71 &4.59& \textbf{11.53} & 91.71 &\textbf{57.33}\\
VGG-19 &Imagenette& 110.35 & \textbf{80.32} & 105.34 & 78.13 &4.54 &\textbf{37.76} & 78.13 &\textbf{65.78}\\
\botrule
\end{tabular*}

\end{minipage}
\end{center}
\end{table*}

\begin{table*}[b!]
\small
\begin{center}
\begin{minipage}{\textwidth}
\caption{Fast Training of ResNet architectures. SGD refers to the standard training strategy of the entire model. Freezing refers to excluding the parameters of some layers from the training without removing the layers from the model. Dropping is our method where layers are deleted from the trained model. T is the training time in minutes. A is the test accuracy value. $\Delta T$ is the percentage of reduced training time with respect to the time of SGD. The table shows that our method allows a reduction of the training time of more than 56\% for all the ResNet architectures trained on different datasets. Similarly to what happens with the VGG model, best accuracy values (in bold) are achieved with SGD. Freezing or Dropping the layers achieve similar accuracy values, which are slightly inferior to the one obtained by SGD. However, in terms of training time reduction $\Delta T$, our method achieves the best results (in bold). 
}
\label{tab:resnet}
\begin{tabular*}{\textwidth}{@{\extracolsep{\fill}}llcccccccc@{\extracolsep{\fill}}}
\toprule%
& &\multicolumn{2}{c@{}@{}}{SGD} & \multicolumn{3}{@{}c@{}}{Freezing}& \multicolumn{3}{@{}c@{}}{Dropping (Ours)} \\
\cmidrule(lr){3-4}\cmidrule(lr){5-7}\cmidrule(lr){8-10}%
\textbf{Network} & \textbf{Dataset} & \textbf{T (min)} & \textbf{A (\%)} & \textbf{T (min)} & \textbf{A (\%)} &\textbf{$\Delta T$ (\%)}& \textbf{T (min)} & \textbf{A (\%)} &\textbf{$\Delta T$ (\%)} \\
\midrule
ResNet-18 &MNIST& 23.67 & \textbf{98.2} & 23.10 & 97.78 &2.41& \textbf{8.64} & 97.78 & \textbf{63.50} \\
ResNet-18 &CIFAR-10& 27.67 & \textbf{92.25} & 25.97 & 91.82 &6.14& \textbf{11.90} & 91.82 & \textbf{56.99}\\
ResNet-18 &Imagenette& 253.07 & \textbf{80.12} & 242.32 & 79.07 &4.25& \textbf{83.78} & 79.07 & \textbf{66.89} \\
\hline
ResNet-50 &MNIST& 35.43 & \textbf{98.75} & 35.02 & 96.85 &1.16&\textbf{ 11.23} & 96.85 & \textbf{68.30}\\
ResNet-50 &CIFAR-10& 38.43 & \textbf{94.40} & 35.40 & 92.05 &7.88& \textbf{13.05} & 92.05 & \textbf{66.04} \\
ResNet-50 &Imagenette& 336.00 & \textbf{82.78} & 315.34 & 80.34 &6.15& \textbf{86.28} & 80.34 & \textbf{74.32} \\
\hline
ResNet-101 &MNIST& 53.12 & \textbf{97.81} & 51.64 & 95.45 &2.79& \textbf{18.56} & 95.45 & \textbf{65.06}\\
ResNet-101 &CIFAR-10& 56.12 & \textbf{93.98} & 52.03 & 91.26 &7.29& \textbf{19.53} & 91.26 & \textbf{65.20}\\
ResNet-101 &Imagenette& 402.34 & \textbf{82.23} & 380.23 & 80.75 &5.29& \textbf{120.44} & 80.75 & \textbf{70.06}\\
\hline
ResNet-152 &MNIST& 70.76 & \textbf{97.43} & 65.30 & 95.12 &7.72& \textbf{23.34} & 95.12 & \textbf{67.01}\\
ResNet-152 &CIFAR-10& 74.76 & \textbf{93.45} & 68.93 & 91.03 &7.80& \textbf{25.75} & 91.03 & \textbf{65.56}\\
ResNet-152 &Imagenette& 540.76 & \textbf{82.65} & 504.72 & 79.45 &6.66& \textbf{180.34} & 79.45 & \textbf{66.65}\\
\botrule
\end{tabular*}

\end{minipage}
\end{center}
\end{table*}

\subsection{Dataset and Hyper-parameters}
To evaluate our algorithm we use three popular classification datasets: MNIST~\cite{lecun1995convolutional}, CIFAR-10~\cite{krizhevsky2009learning}, and Imagenette~\cite{Imagenette}.

The MNIST dataset contains 60,000 gray scale images, with 50,000 for training and 10,000 for test. It includes 10 classes, and each class is represented by 6,000 images.

The CIFAR-10 dataset contains 60,000 color images, with 50,000 for training and 10,000 for test. Overall, there are 10 classes, and, for each class, 6,000 images.

Imagenette \cite{Imagenette} includes a subset of 10  classes from the larger Imagenet~\cite{russakovsky2015imagenet}. 

The classes are {\em tench, English springer, cassette player, chain saw, church, French horn, garbage truck, gas pump, golf ball, parachute}. Overall, the adopted dataset includes about 1,000 color images per class with a resolution of 160 x 160 pixels. The images are obtained from the ones in the original Imagenet dataset by performing a resizing that preserves the original aspect ratio.

We opted to use MNIST and CIFAR-10 datasets to ensure the comparability of our work with~\cite{xiao2019fast}. In addition, we included the more complex Imagenette dataset to evaluate the performance of our work on a more challenging dataset.

In all our experiments, we use a batch size of 256 samples.
We set the learning rates to values generally used in literature, while the number of epochs and warm up are selected empirically. On the MNIST and CIFAR-10 datasets, the total number of epochs (including the model warm-up) is set to 60.
On the MNIST dataset, the learning rate is fixed to 0.001. The warm-up epochs are 5. On the CIFAR-10 dataset, the learning rate is fixed to 0.1 and scaled x10 after 20 epochs. The warm-up epochs are 10. Finally, on the Imagenette, the number of epochs is set to 150 and the learning rate is fixed to 0.01 and scaled x10 after 50 epochs. The warm-up epochs are 25.

\begin{figure*}[t!]
\centering
\includegraphics[width=.49\linewidth]{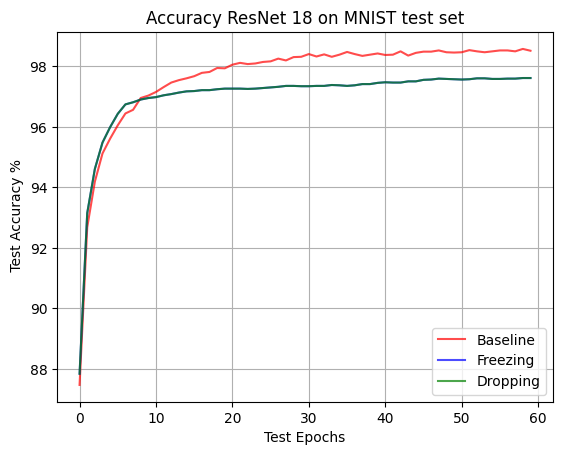}
  \includegraphics[width=.49\linewidth]{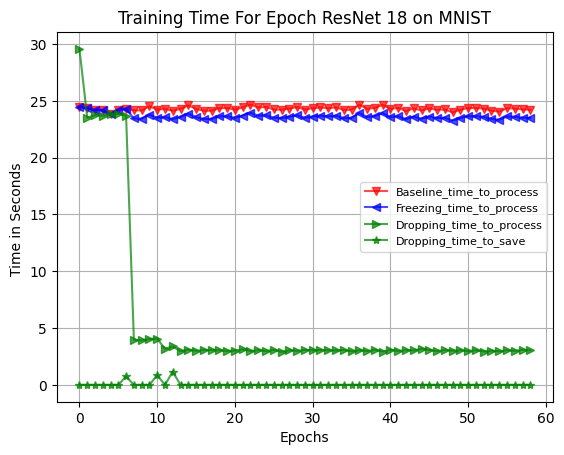}
  \includegraphics[width=.49\linewidth]{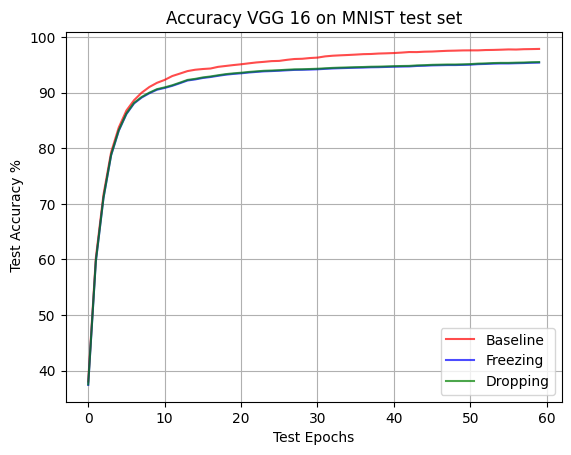}
  \includegraphics[width=.49\linewidth]{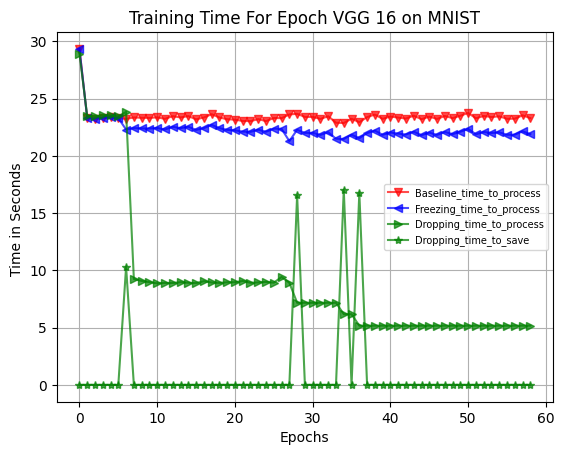}
  
  \caption{The plots on the left show the test accuracy values of a ResNet-18 (top) and VGG-16 (bottom) trained on the MNIST dataset with different strategies: SGD (red curves), layer freezing (blue curves), and layer dropping (green curves). The experiments were repeated 10 times with different starting weights and data randomization. Freezing and dropping layers achieve nearly equivalent test accuracy values for both the architectures. The values are slightly lower than those achieved when training the entire model. On the right, the plots show training time per epoch for the same models and dataset. The red and blue curves represent the training time for the SGD and layer freezing strategies respectively. The green curves refer to our layer dropping approach. Starred curves show the time required to store the feature maps to disk, while the other curves show the training time which decreases over the epochs due to the lower cost of forward propagation in our method. Note that Intermediate feature maps in VGG-16 have greater size than in ResNet and the time to store the maps to disk is higher. However, this operation is not frequent and its cost is amortized over time.
  }
  \label{fig:acc_time_res}
\end{figure*}

\begin{figure*}[b!]
\centering
\includegraphics[width=.49\linewidth]{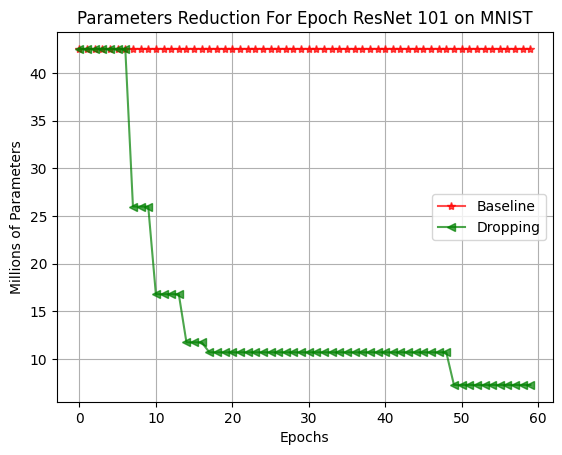}
  \includegraphics[width=.49\linewidth]{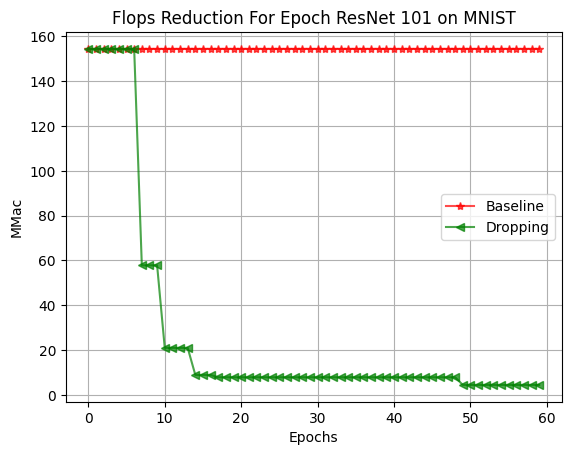}
    \includegraphics[width=.49\linewidth]{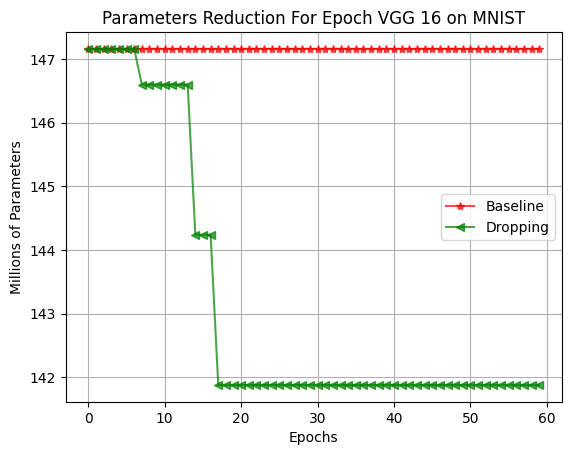}
  \includegraphics[width=.49\linewidth]{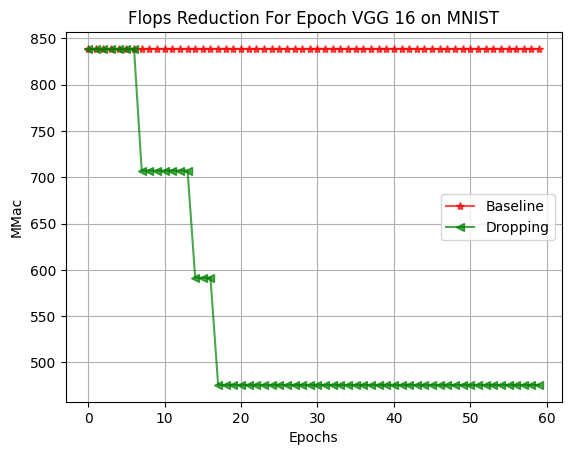}

  \caption{On the left, the plot shows the number of network parameters in each epoch for the ResNet-101 (top) and VGG-16 (bottom). When training the entire model or using the layer freezing technique (red curves), the number of parameters remains constant. With the proposed method (green curves), the model shrinks over the epochs and there is a reduction in the number of parameters each time a sequence of layers is dropped. This parameter reduction is correlated with the MMac reduction, shown in the plots on the right for both the models. By reducing the network parameters, the number of operations during forward propagation is decreased.}
  \label{fig:par_flops_res}
\end{figure*}

\subsection{Results and Comparison}

Table~\ref{tab:vgg} reports the results obtained by applying our method to models of various depth (11, 16, and 19) in the family of VGG architectures. All models include batch-normalization. First we note that, across the models and datasets, the impact on accuracy values of freezing or dropping  layers to increase training efficiency is negligible. The differences in the accuracy values vary, for both techniques, in the range $0.26$ (for VGG-16 trained on CIFAR-10) and $2.38$ (for VGG-19 trained on MNIST). These differences increase slightly with network depth and are generally higher for the Imagenette dataset. However, these small decreases in accuracy values come with a reduction in training time.
As shown in the table, for VGG models, while the training time reduction with the freezing layer technique varies in the range $0.40\%$ (for VGG-16 on the MNIST dataset) and $6.03\%$ (for VGG-16 on the CIFAR-10 dataset), with our layer dropping technique the training time reduction varies in the range of $58.04\%$ (for VGG-11 on the MNIST dataset) and $69.97\%$ (for VGG-11 on the Imagenette dataset). While on average these percentages are $3.52\%$ for the freezing layer method, they are $62.87\%$ with our technique. 

These results are not limited to VGG architectures. Indeed, Table~\ref{tab:resnet} reports similar achievements for ResNet architectures of various depth (18, 50, 101, and 152). In particular, analyzing the results in a similar way to what was done for the VGG architectures, the differences in the accuracy values vary, for the layer freezing and dropping techniques, in the range $0.4\%$ (for ResNet-18 on the MNIST dataset) and $3.2\%$ (for ResNet-152 on the Imagenette dataset). For both techniques, the impact on accuracy values generally increases with network depth.
In terms of training time reduction, with the layer freezing technique, the percentages vary in the range $1.16\%$ (for ResNet-50 on the MNIST dataset) and $7.8\%$ (for Resnet-152 on the CIFAR-10 dataset). Our approach achieves training time reduction percentages in the range $56.99\%$ (for ResNet-18 on the CIFAR-10 dataset) and $74.32\%$  (for ResNet-50 on the Imagenette dataset). While, on average, layer freezing accounts for a $5.46\%$ of training time reduction, our method results in average training time reduction of approximately $66.30\%$. 

Overall, the training time for the VGG and ResNet architectures is more than halved with our technique despite the loss in accuracy values being comparable to that obtained by freezing the layers.

\begin{table}[t!]
\begin{center}
\caption{FLOPs reduction across architectures. SGD refers to the standard training strategy of the entire model. Dropping is our method where layers are deleted from the trained model. FLOPs are measured during the forward propagation. DeltaFLOPs is the percentage of reduced FLOPs with respect to SGD. As shown in bold, our approach reduces the FLOPs for all architectures, and the reduction is higher for larger models. Thus, our method improves training time not only because it reduces the number of gradients to be computed in  back-propagation but also because it reduces the FLOPs in the forward propagation.
}
\label{tab:flops}
\begin{tabular}
{@{\extracolsep{\fill}}lccc@{\extracolsep{\fill}}}
\toprule%
& \multicolumn{1}{c@{}}{SGD} &\multicolumn{2}{@{}c@{}}{Dropping (Ours)} \\
\cmidrule(lr){2-2}\cmidrule(lr){3-4}%
\textbf{Network} & \textbf{FLOPs} & \textbf{FLOPs} &\textbf{$\Delta$FLOPs ($\%$)}\\
\midrule
%\hline
VGG-11  & 31,203.60 & \textbf{25,847.02} & 17.17 \\
VGG-16  & 50,311.19 & \textbf{33,049.44} & 34.31 \\
VGG-19  & 66,000.00 & \textbf{44,079.31} & 33.21 \\
\hline
ResNet-18 & 1,987.80 & \textbf{640.53} & 67.78 \\
ResNet-50 & 4,704.59 & \textbf{2,193.45} & 53.38 \\
ResNet-101 & 9,262.2 & \textbf{1,667.27} & 82.00 \\
ResNet-152 & 13,823.39 & \textbf{2,247.46} & 83.74 \\
\botrule
\end{tabular}
\end{center}
\end{table}

\section{Training Time and Parameter Reduction}
\label{sec:tr}

The results discussed in Sec.~\ref{sec:exp} refer to the impact of our technique on model accuracy and the training time reduction achieved on our machine.
To further analyze and explain the performance of our technique, this section refers to the effect it has on the number of parameters and operations performed during forward propagation at training time. 
Indeed, our technique not only reduces the number of weights for which it is necessary to estimate partial derivatives during gradient back-propagation, but also affects the number of operations performed during forward propagation.

A potential bottleneck in our method is the feature map saving to disk whenever layer dropping occurs.
In some networks, the size of the features maps produced by a layer may be greater than the size of the input images; thus, reading and writing feature maps with large dimensions can cause a slowdown of the training. However, we have observed empirically that layers dropping never takes place layer by layer but generally several subsequent layers are dropped together.
Figure~\ref{fig:acc_time_res} shows (on the left) the test accuracy values of ResNet-18 (top row) and VGG-16 (bottom row) on the MNIST dataset over the epochs for SGD, the layer freezing method and our technique. As already discussed, our technique and the freezing layers one have a limited impact on the final model accuracy. On the right, the figure shows the training time per epoch of each technique. In particular, the time spent by our method refers to: the time to store feature maps on the disk whenever layer dropping arises, and the time to train the model. As shown in the plot, the time to store feature maps on the drive is concentrated only in few epochs since, as already said, our method does not constantly drop layers. The time to store the maps to disk depends on their size and, thus, is higher for the VGG model. The time to train the model includes, for all methods, the time for computing the gradients, updating the layer weights, loading the training data and performing forward propagation. The plot shows how, in our technique, this time decreases over time and becomes much lower than that of the other methods. 

To further investigate this result, we discuss the FLOPs (floating point operations per second) of our method versus the baseline method (SGD) and the layer freezing technique we are comparing. More specifically, we measured the MMAC (Mega Multiply-Accumulate), a FLOPs metric that counts the number of matrix multiplications and accumulations (MACs) a neural network performs in one second. The metric is expressed in millions (mega) of operations and is useful for evaluating the computational complexity of a neural network and comparing the performance of different architectures.
Fig.~\ref{fig:par_flops_res} shows, on the left, the number of parameters per each epoch for the ResNet-101 (top) and VGG-16 (bottom). Red curves represent the number of model parameters when using the baseline and the layer freezing techniques; Green curves  are the number of parameters when using our method. Since our method compresses the model at training time, there is a strong reduction in the number of parameters corresponding to the epochs when layer dropping arises. This reduction of parameters is correlated with the MMAC reduction during forward propagation, shown in the plots on the right. While the MMAC is constant for the baseline and the layer freezing techniques, in our method it keeps decreasing due to the shrinkage in the number of parameters.

Table~\ref{tab:flops} reports the FLOPS required for the forward propagation during training by using SGD versus our approach (Dropping). The FLOPs refers to various deep learning models, including VGG-11, VGG-16, VGG-19, ResNet-18, ResNet-50, ResNet-101, and ResNet-152.
The second column shows the FLOPs required when training the entire model, the third column shows the FLOPs required when using our approach. The final column $\Delta$FLOPs shows the percentage difference between the FLOPs  of the two approaches.

Overall, the table suggests that the Dropping approach is more efficient than the baseline method. The percentage difference between the two approaches ranges from 17.17\% (for VGG-11) to 83.74\% (for ResNet-152), indicating that Dropping can significantly reduce the computational burden of training deep neural networks, especially for very deep models like ResNet-101 and ResNet-152.
The table also shows that the reduction of the FLOPs increase with the depth of  the model. Moreover, has also indicated by the magnitude of  the FLOPs, the VGG family performs more operations than the ResNet  due to the presence of dense layers applied to feature maps of greater size.

\section{Conclusion and Future Works}
\label{sec:con}
In this work, we have proposed a method to improve the training efficiency of convolutional networks by gradually compressing the model through dropping of convolutional layers during the training phase. While most of the state-of-the-art methods focus only on reducing time and operations during back propagation, we have noticed that forward propagation contributes significantly to the overall computational cost of the training procedure.

We have empirically found that in the ResNet and VGG architectures, the layers closest to the input learn faster than those close to the output. Therefore, the proposed method uses a gradient-based score to identify the layers to be dropped. Dropping arises from the first layer of the network to the last.
In contrast to previous work, which freeze layer parameters to speed up model training, our approach is to remove layers from the network only during training. To this end, our method splits the network into a ``tail", consisting of the dropped layers, and a ``head", consisting of the layers yet to be trained. The tail is used to prepare the data to feed the head. The head is a smaller version of the model that trains over time.

The method has been validated on three popular datasets to train models of various depth in the VGG and ResNet families. Our experiments show that, on average, our method achieves a time reduction of $62.87\%$ on VGG architectures, and of $66.30\%$ on ResNet models. Thus, by reducing the time of forward propagation during training, the overall training time is more than halved with a limited impact on the model accuracy that, in our experiments, never exceeds the $3.2\%$ for very deep network (ResNet-152).

A potential limitation of our method is that it requires a warm-up training to initialize the network weights. This is often done in fast training algorithms and network compression  to avoid pruning important weights. In these works, as in ours, the number of warm-up epochs is found empirically. However, in our method, it might be equivalently possible to define a threshold on the layer score to start dropping layers when their weights no longer change much. Since our method significantly reduces the training time, it might also be possible to use a hybrid strategy that alternates among training the whole model and the head model. This strategy can help refine the first dropped layers' weights at the cost of increasing the overall training time.

Theoretically, the method can also be used on dense layers. However, in our experiments, we have observed that the layer scores of dense layers do not always have increasing values. Thus, there is a risk of dropping a dense layer whose weights can still improve. 

In future work, we intend to investigate to what extent our approach can be applied to other more complex architectures such as multi-branch models like Siamese networks or models designed for time series processing, such as those  that include recurrent memory cells (Recurrent Neural Network - RNN). Indeed, in multi-branch networks one should check whether learning proceeds similarly along each of the branches, which is not obvious considering that each branch can process a different type of data. In case of recurrent memory cells, the method may be applied to convolutional layers preceding the cells, which can increase the time needed to store the feature maps to disk, since it has to be done for each temporal sample, but it can also be interesting to study how the gradient-based scores of these memory cells change over the epochs. Furthermore, it will be useful to study whether similar techniques can be applied to visual transformers.

Considering that our approach allows us to reduce the model size with positive consequences on the memory usage, we will investigate an adaptive method to estimate optimal batch size while layer dropping is applied. Indeed, as the number of layers in the model is reduced over time, the overall complexity of the model decreases and also the memory usage for feature maps and gradients. Thus, since the training would use less memory, we may dynamically increase the batch size. This approach would reduces the number of iterations in an epoch and probably improve the efficiency of our model.

\bibliography{sn-bibliography}
\end{document}